\documentclass[10pt,journal,compsoc]{IEEEtran}
\usepackage{cite}
\usepackage{multirow}
\usepackage{graphicx}
\usepackage{color}
\usepackage{dcolumn}
\usepackage{amsmath}
\usepackage{kantlipsum}
\usepackage{url}
\usepackage{microtype}
\usepackage{verbatim}
\usepackage{colortbl}
\usepackage{arydshln}
\usepackage{multirow}
\newcolumntype{K}[1]{>{\centering\arraybackslash}p{#1}}
\usepackage{booktabs}
\usepackage[flushleft]{threeparttable}
\usepackage{hhline}
\usepackage{algorithm}
\usepackage{algorithmic}
\usepackage{mathtools}
\usepackage{breqn}

\ifCLASSOPTIONcompsoc
\usepackage[caption=false, font=normalsize, labelfont=sf, textfont=sf]{subfig}
\else
\usepackage[caption=false, font=footnotesize]{subfig}
\fi
\begin{document}
\title{MHDeep: Mental Health Disorder Detection System based on Body-Area and Deep Neural Networks}

\author{Shayan~Hassantabar,~Joe~Zhang,~Hongxu Yin,
and~Niraj~K.~Jha,~\IEEEmembership{Fellow,~IEEE}% <-this % stops a space
\thanks{
This work was supported by NSF under Grant No. CNS-1907381.  Shayan Hassantabar, Joe Zhang,
Hongxu Yin, and Niraj K. Jha are with the Department of Electrical Engineering, Princeton
University, Princeton, NJ, 08544 USA, e-mail:\{seyedh,zhaoz,hongxuy,jha\}@princeton.edu.}}

\IEEEtitleabstractindextext{%
\begin{abstract}
Mental health problems impact quality of life of millions of people around the world.
However, diagnosis of mental health disorders is a challenging problem that often relies on
self-reporting by patients about their behavioral patterns and social interactions.
Therefore, there is a need for new strategies for diagnosis and daily monitoring of mental health
conditions. The recent introduction of body-area networks consisting of a plethora of accurate
sensors embedded in smartwatches and smartphones and edge-compatible deep neural networks (DNNs) points
towards a possible solution. Such wearable medical sensors (WMSs) enable continuous monitoring of
physiological signals in a passive and non-invasive manner.  However, disease diagnosis based on WMSs
and DNNs, and their deployment on edge devices, such as smartphones, remains a challenging problem.
To this end, we propose a framework called MHDeep that utilizes commercially available WMSs and
efficient DNN models to diagnose three important mental health disorders: schizoaffective, major
depressive, and bipolar.  MHDeep uses eight different categories of data obtained from sensors
integrated in a smartwatch and smartphone. These categories include various physiological signals and
additional information on motion patterns and environmental variables related to the wearer.
MHDeep eliminates the need for manual feature engineering by directly operating on the data streams
obtained from participants. Since the amount of the data is limited, MHDeep uses a synthetic data
generation module to augment real data with synthetic data drawn from the same probability
distribution.  We use the synthetic dataset to pre-train the weights of the DNN models, thus
imposing a prior on the weights.  We use a grow-and-prune DNN synthesis approach to learn both the
architecture and weights during the training process.  We use three different data partitions to
evaluate the MHDeep models trained with data collected from 74 individuals.  We conduct two types of 
evaluations: at the data instance level and at the patient level. MHDeep achieves an average test 
accuracy, across the three data partitions, of 90.4\%, 87.3\%, and 82.4\%, respectively, for 
classifications between healthy and schizoaffective disorder instances, healthy and major depressive 
disorder instances, and healthy and bipolar disorder instances.  At the patient level, MHDeep DNNs 
achieve an accuracy of 100\%, 100\%, and 90.0\% for the three mental health disorders, 
respectively, based on inference that uses 40, 16, and 22 minutes of data from each patient.
\end{abstract}

% Note that keywords are not normally used for peerreview papers.
}
\maketitle

\IEEEdisplaynontitleabstractindextext

\IEEEpeerreviewmaketitle

\IEEEraisesectionheading{\section{Introduction}\label{sect:introduction}}

\IEEEPARstart{M}ental health problems impact around $20\%$ of the world population
\cite{steel2014global}.  They may negatively affect a person's mind, emotions, behavior,
and even physical health. Mental health issues may include various disorders like bipolar,
depression, schizophrenia, and attention-deficit hyperactivity, to name but a few.  These
disorders do not only affect adults, but children and adolescents may suffer
from them as well \cite{polanczyk2015annual}.  Moreover, patients with serious mental health issues
have a higher risk of morbidity due to physical health problems. Depression, for example, can lead to
disability and increases the risk of suicidal thoughts and attempts.

In order to understand the mental health condition of the patient and provide suitable patient care,
early detection is essential. However, this remains a public health challenge.  While many other
diseases can be diagnosed based on specific medical tests and laboratory measurements, detection of
mental health problems mainly relies on self-reports and responses to specific questionnaires
designed for identifying certain patterns of behavior and social interactions.
Hence, to address this challenge, novel detection strategies are needed.

There has been recent interest in employing machine learning (ML) to detect mental health
conditions \cite{dwyer2018machine}. Neural networks are popular machine
learning models that use nonlinear computations to make inferences from large datasets. Thus,
they have started being deployed in the smart healthcare domain \cite{hassantabar2020coviddeep,
hassantabar2020diagnosis, miotto2018deep, yin2019diabdeep, akmandor2017smart, yin2018smart}.

In previous studies, two main data sources for deep learning based analysis of mental health
have been clinical data and social media usage data. The former includes studies that use
neuro-image data for detecting various mental health disorders \cite{schnack2014can}, electroencephalogram (EEG) data to study brain disorders
\cite{acharya2018automated}, and analysis of electronic health records (EHR)
to study mental health problems \cite{geraci2017applying}.  Moreover, social media usage patterns
have been used to predict personal traits of the user. As a result, several recent works focus on
exploiting such patterns to detect psychiatric illness \cite{birnbaum2020identifying}.

\begin{figure}[!ht]
    \centering
    \includegraphics[scale= 0.5]{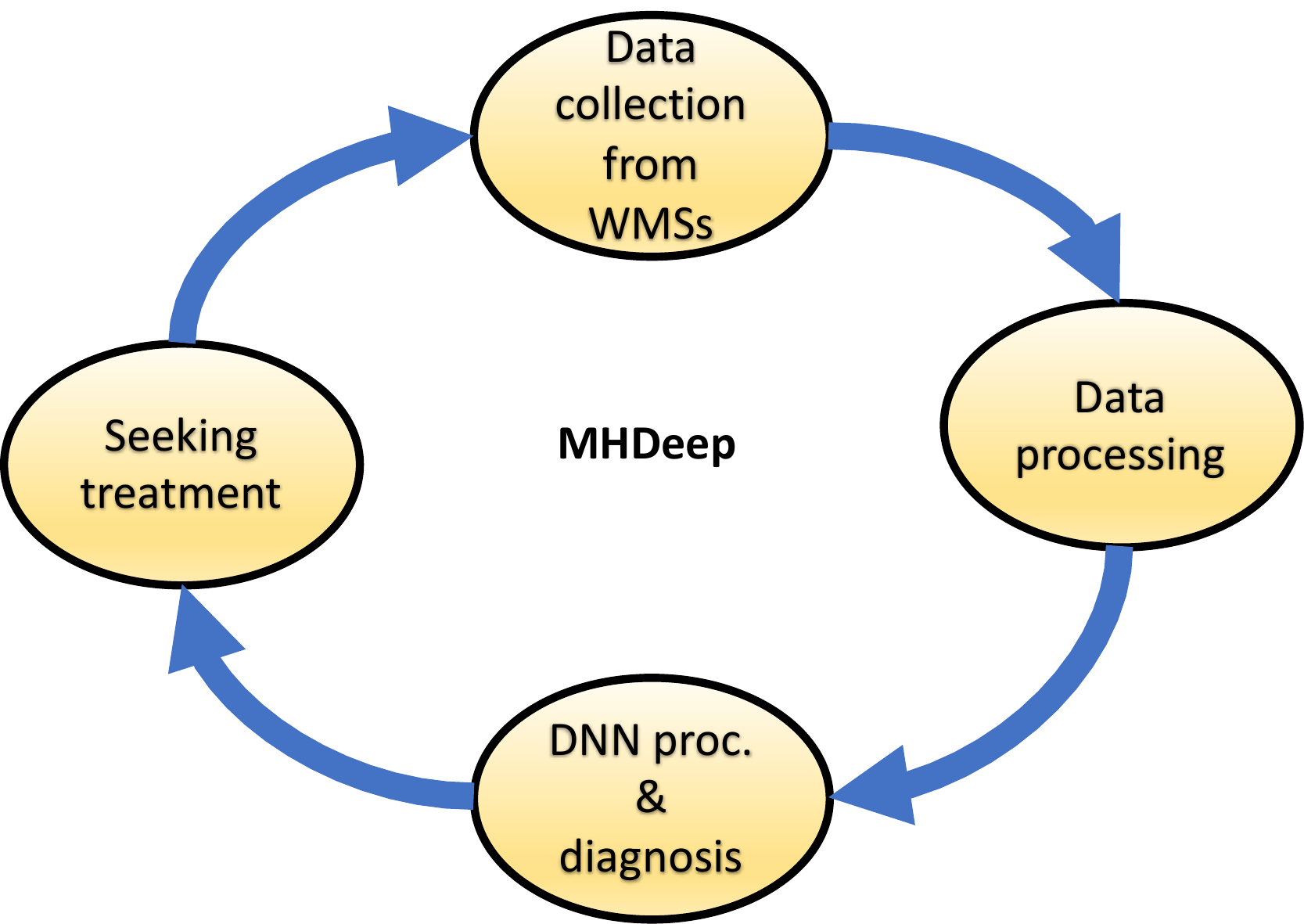}
    \caption{MHDeep mental health disorder detection system}
\label{fig:MHDeep}
\end{figure}

Although the above works have demonstrated the promise of using machine learning in identifying
mental health disorders, daily mental health monitoring is still a challenge.  Since mental
health condition treatment delays may lead to negative outcomes, potentially even loss of life, it
is desirable to have immediate and pervasive mental health detection.  This is the motivation behind
our mental health detection system, MHDeep.  As shown in Fig.~\ref{fig:MHDeep}, MHDeep relies on
physiological data collected using various WMSs.  WMSs can be used to continuously monitor the
physiological signals of the wearer throughout the day. This enables constant tracking of the health
conditions of the user.  MHDeep uses various sensors embedded in smartwatches and smartphones.
For training purposes, the collected physiological data are processed to obtain a comprehensive
dataset.  MHDeep combines data from WMSs with the inference capabilities of deep neural networks
(DNNs) to directly extract mental health condition from the physiological signals.  These
inferences can be communicated to a health server that is accessible to the physician.  This has the
potential to enhance to ability of the physician to intervene quickly when mental health
conditions deteriorate.

Difficulty of data collection and labeling limits the amount of available data. Hence,
MHDeep uses synthetic data drawn from the same probability distribution as real data to augment
the dataset.  It also leverages a grow-and-prune DNN synthesis approach \cite{dai2019nest, hassantabar2019scann} to train accurate and
computationally efficient neural network models to detect the mental health condition of the user.

The major contributions of this article are summarized next.
\begin{itemize}
    \item We demonstrate an easy-to-use, accurate, and pervasive mental health disorder detection 
system, called MHDeep. MHDeep combines physiological signals collected from WMSs with the prediction 
power of DNNs to detect three main mental health disorders: major depressive, schizoaffective, 
and bipolar.  Unlike many other approaches for detecting mental health problems, MHDeep does not 
rely on any self-reports from the user. 
    \item We do an extensive search to extract the most appropriate set of data categories for 
detecting each of the three mental health disorders. 
    \item MHdeep relies on a synthetic data generation module to alleviate the concerns arising
from unavailability of large datasets.  It uses a grow-and-prune DNN synthesis approach to improve 
the accuracy of the DNNs while reducing their computational costs. 
    \item We demonstrate the performance, accuracy, and feasibility of MHDeep through extensive 
evaluations.
\end{itemize}

The rest of the article is organized as follows.
Section \ref{sect:backgroung} presents background information on various works related to MHDeep. 
Section \ref{sect:methodology} explains the MHDeep framework thoroughly. 
Section \ref{sect:implementation} provides implementation details. Section \ref{sect:evaluation}
presents experimental evaluations. Section \ref{sect:discussion} provides a short discussion related 
to this work.  Finally, Section \ref{sect:conclusion} concludes the article. 

\section{Background}
\label{sect:backgroung}
In this section, we first provide background information on various mental health disorders and how 
they affect patient lives. Next, we discuss various methods for identifying mental health conditions 
based on machine learning. We also discuss some of the related work on synthesizing
efficient neural network models. Finally, we discuss WMSs and their applications to various disease 
diagnosis frameworks. 

\subsection{Mental health disorders and their impact}
Mental health conditions can affect a patient's thinking, feeling, and behavior. They may have a deep
impact on the daily life of the person and affect their ability to adequately perform in society. 
There are hundreds of different mental illnesses \cite{lehman2000diagnostic}. We discuss the three 
mental health disorders that we target in this work: bipolar, major depressive, and schizoaffective. 

Bipolar disorder can cause a dramatic shift in a person's mood, energy, and behavior. It is 
characterized by experiences of alternating episodes of manic and depressed states. 
Major depressive disorder may present different symptoms like loss of interest, sleep disturbance, 
change in appetite, and feeling of fatigue.  Schizoaffective disorder is characterized by various 
symptoms of schizophrenia such as episodes of hallucinations and delusions. It may also present other 
symptoms such as disorganized thinking, depressed mood, and manic behavior. 

Apart from various conditions that these mental health disorders may cause, stereotypes related to 
mental health seem to still be widely prevalent in society, not just among uninformed people but even
among well-trained professionals \cite{corrigan2000mental}.  These stereotypes often lead to social 
and employment discrimination \cite{bordieri1986hiring} and poor treatment of physical health 
problems \cite{corrigan2014mental}. 

\subsection{Deep learning for mental health}
Deep learning has been recently used to better understand and detect mental health problems. 
Deep learning approaches have been applied to various types of data: mainly clinical data and 
social media usage data \cite{su2020deep}.  The three types of clinical data used in these works are 
neuroimage data, EEG data, and EHR data. 

Several studies demonstrate the effectiveness of neuroimages in detecting neuropsychiatric disorders 
\cite{schnack2014can}. Two types of neuroimage data used in such works are functional magnetic 
resonance imaging (fMRI) and structural magnetic resonance imaging (sMRI).  fMRI measures brain 
activity by monitoring blood oxygenation and flow in response to neural activity. sMRI examines the 
anatomy and pathology of the brain.  Deep belief networks have been used to detect the presence of 
attention-deficit hyperactivity disorder (ADHD) using fMRI and sMRI data 
\cite{kuang2014classification, kuang2014discrimination}.  These data types have also been used to 
detect schizophrenia \cite{zeng2018multi, pinaya2016using}.  Depression has been detected using 
time-series fMRI data using convolutional neural networks (CNNs) and autoencoders 
\cite{xiang2017application}.  EEG is another source of data for studying brain disorders. For 
example, CNN based feature extraction from EEG data has been used to detect depression 
\cite{acharya2018automated}. EHR is a collection of patient-centered records and 
includes both structural data such as laboratory reports and unstructured data such as clinical and 
discharge notes.  Since EHR data is a collection of longitudinal records, recurrent neural networks 
(RNNs) have been used to distill information from them.  Pham et al.~\cite{pham2017predicting} use 
RNN architectures to predict future outcomes of depressive episodes.  Unstructured clinical notes 
have been analyzed with deep learning-based models to detect depression \cite{geraci2017applying}.  

Social media usage data have also proved their usefulness in identifying psychiatric illness. 
Birnbaum et al.~\cite{birnbaum2020identifying} investigate Facebook messages and patterns of sharing 
images on social media to distinguish among healthy individuals, individuals with a schizophrenia 
spectrum disorder, and individuals with mood disorders.  Other works have used DNNs with textual 
data and image data shared on social media platforms to detect stress \cite{lin2014user}, depression 
\cite{sadeque2017uarizona}, and risk of suicide \cite{coppersmith2018natural}. 

\subsection{Efficient neural network synthesis}
Next, we summarize the main approaches to synthesis of compact DNN models. Conventional 
synthesis methods are based on the use of efficient building blocks. For example, MobileNetV2 
\cite{sandler2018mobilenetv2} leverages inverted residual blocks to reduce model size and 
computations significantly. Wu et al.~\cite{wu2018shift} use shift-based operations rather than 
convolution layers to significantly reduce computational costs of the model. The main 
drawback of such approaches is the need for considerable design insight and trial-and-error process 
to design such efficient building blocks.  Network compression is another approach to design of 
efficient models.  It removes the need for design insights.  Network pruning is a widely used method 
that eliminates weights or filters that do not enhance model performance. Han et 
al.~\cite{han2015deep} have shown the effectiveness of pruning in removing redundancy in CNNs and 
multilayer-perceptron architectures.  Grow-and-prune DNN synthesis uses network growth
followed by network pruning in an iterative process to improve model performance 
while ensuring its compactness \cite{hassantabar2019scann, dai2019nest}.  

Another recent approach relies on the use of reinforcement learning (RL) to search for DNN 
architectures in an automated flow. It is known as neural architecture search (NAS) 
\cite{DBLP:conf/iclr/ZophL17}. NAS generally uses a controller, e.g., 
an RNN, to iteratively generate candidate architectures in the search process. The RL
controller is improved based on candidate performance. As an example, MnasNet \cite{tan2019mnasnet} 
uses an RL-based approach to develop efficient DNNs for mobile platforms. However, the downside of 
the RL-based NAS approach is that it is computationally intensive. FBNet \cite{wu2019fbnet} uses the 
Gumbel softmax function to optimize weights and connections using a single objective function. 
NEAT \cite{stanley2002evolving} uses evolutionary algorithms to generate optimized and increasingly 
complex architectures over multiple generations. Combining efficient evolutionary search algorithms 
with various performance predictors, e.g., for accuracy, energy, and latency, is another approach 
for synthesizing accurate yet compact CNNs/DNNs \cite{dai2018chamnet, hassantabar2019steerage}. 

\subsection{Wearable medical sensors}
Due to recent developments in low-power sensor design and efficient wireless communications, 
battery-powered WMSs are becoming ubiquitous.  More than 123 million WMSs were sold worldwide in 
2018. This number is projected to grow to 1 billion by the end of 2022 \cite{wms-sold}. 
WMSs can track different aspects of human health including heart rate, body/skin temperature, 
respiration rate, blood pressure, EEG, electrocardiogram (ECG), and Galvanic skin response (GSR) 
\cite{baig2013smart}.  Furthermore, the number of physiological signals that can be measured using 
WMSs keeps growing every year. 

WMSs have begun to be used in many smart healthcare applications. CodeBlue \cite{malan2004codeblue} 
is a sensor network that collects vital health signs and transmits them to the healthcare provider. 
MobiHealth \cite{wac2006qos} is a WMS based body-area network (BAN) that realizes an end-to-end mobile 
health monitoring platform. Yin et al.~\cite{yin2019diabdeep} use WMSs for pervasive diagnosis of 
Type-I and Type-II diabetes. CovidDeep \cite{hassantabar2020coviddeep} is a WMS-based framework for 
quick detection of SARS-CoV-2/COVID-19. 

For data collection in MHDeep, we use an Empatica E4 smartwatch \cite{e4-connect} to record a 
subset of patient's physiological signals.  It is a wearable wireless device designed for 
comfortable, continuous, and real-time data acquisition.  We also use a smartphone to simultaneously 
record signals related to motion information and environmental variables. 
Since the DNNs developed for diagnosing various mental health conditions can
reside on the smartphone, use of a smartwatch/smartphone based BAN can enable accurate, yet
convenient, disease diagnosis and continuous healthcare monitoring.

\section{Methodology}
\label{sect:methodology}
In this section we describe various parts of the MHDeep framework.  First, we give an overview of our 
approach.  Then, we discuss the data collection and preparation process, synthetic data generation, 
and grow-and-prune DNN synthesis. 

\subsection{The MHDeep framework}
\begin{figure*}[!ht]
    \centering
    \includegraphics[scale=0.5]{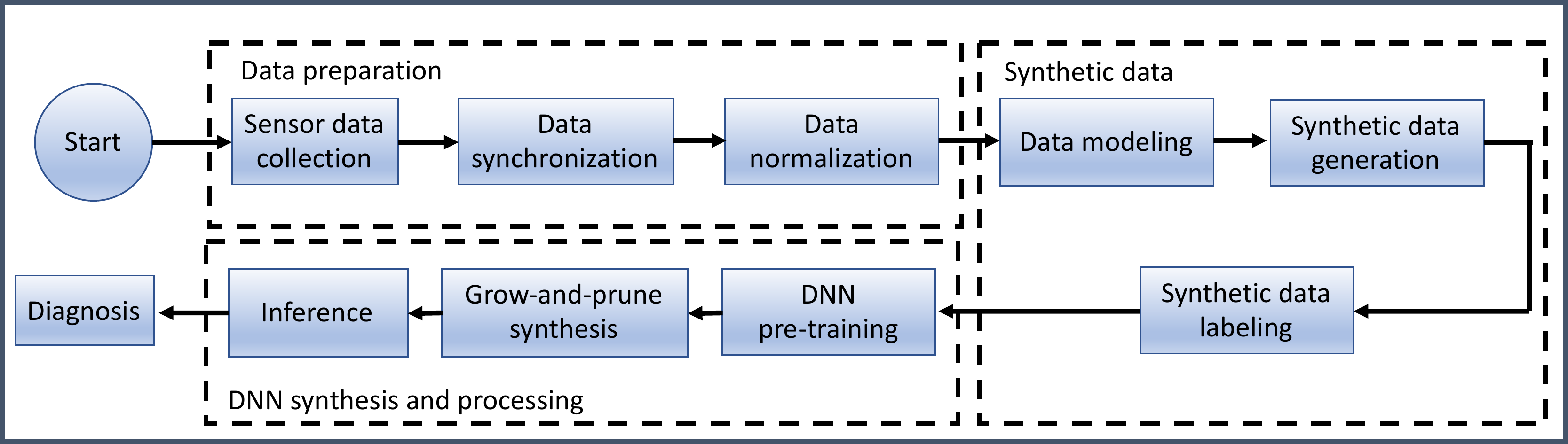}
    \caption{Schematic diagram of the MHDeep framework.}
\label{fig:diagram}
\end{figure*}

We illustrate the MHDeep framework in Fig.~\ref{fig:diagram}. The input data are derived from the 
physiological signals collected using various WMSs in the smartwatch and smartphone in a non-invasive, 
passive, and efficient manner. The list of collected data streams include GSR, skin temperature
(ST), inter-beat interval (IBI), and 3-way acceleration from the smartwatch. 
In addition, some information related to motion patterns of the user and ambient information 
are collected using smartphone sensors.  This includes ambient temperature, gravity, acceleration, 
and angular velocity.  After sensor data collection, the collected signals are synchronized, 
aggregated, and merged into a comprehensive data input for subsequent analysis.  To enhance the 
accuracy of subsequent analysis and improve noise tolerance, we normalize the data. The process of 
data collection and preparation is discussed in more detail in Section~\ref{sect:data-collection}. 
When the size of the training dataset is small, it can be useful to generate a synthetic dataset from 
the same probability distribution as the real training dataset. MHDeep leverages Gaussian mixture 
model (GMM) based density estimation to generate the synthetic data.  Then, it uses grow-and-prune DNN 
synthesis to generate inference models that are both accurate and computationally efficient. 
Section~\ref{sect:dnn-synthesis} discusses the MHDeep DNN synthesis process in detail. 
MHDeep generates DNN architectures that are efficient enough to be deployed on the edge devices such 
as smartphones or smartwatches. Section~\ref{sect:inference} discusses the inference process of the 
MHDeep DNNs for diagnosis and daily monitoring of mental health issues. 

\subsection{Data collection and preparation}
\label{sect:data-collection}
\begin{figure}[!ht]
    \centering
    \includegraphics[scale= 0.5]{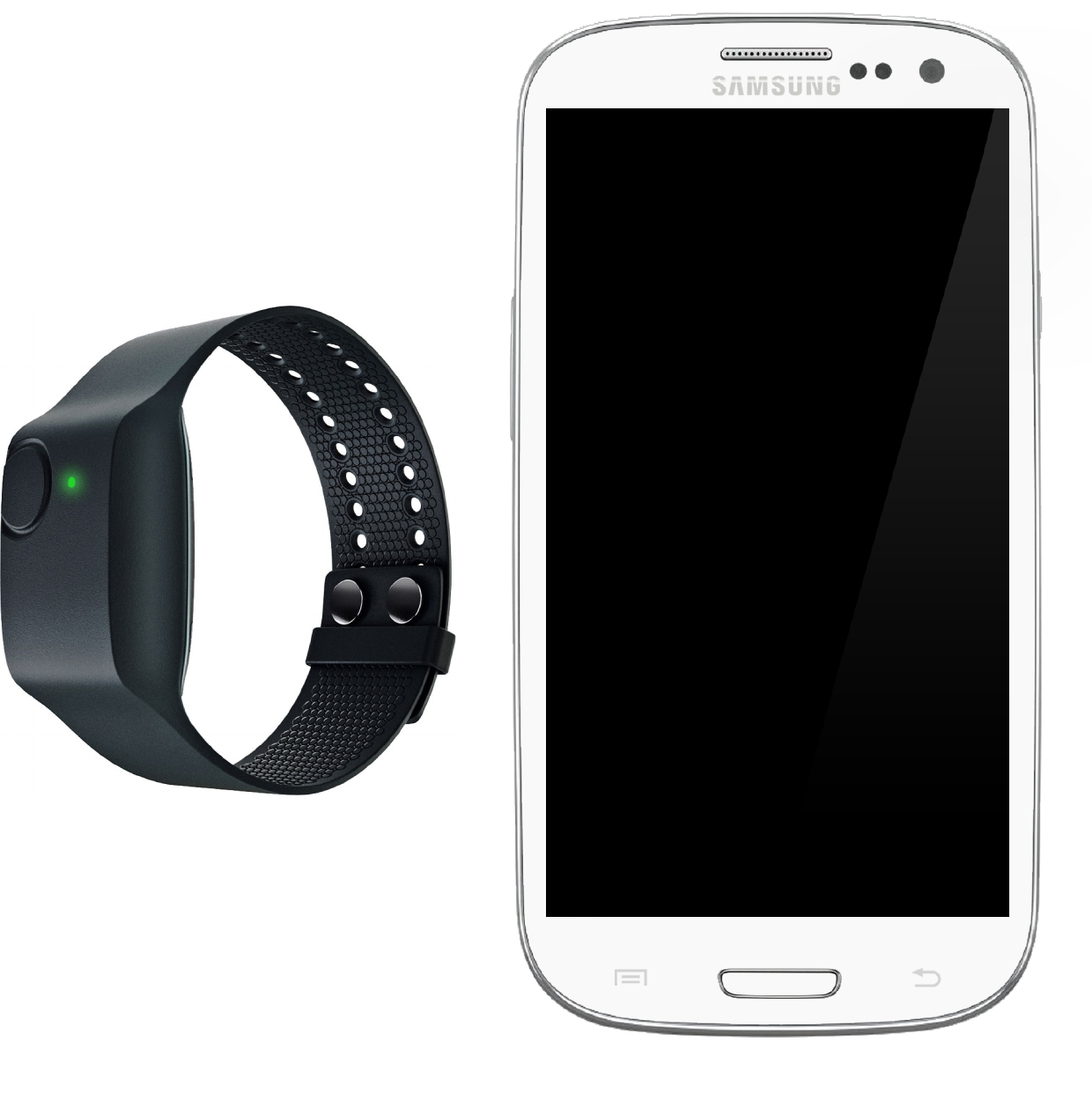}
    \caption{An Empatica E4 smartwatch (left) and Samsung Galaxy S4 smartphone (right) used in the 
data collection process.}
\label{fig:devices}
\end{figure}

We collected WMS data from a total of 74 adult participants at the Hackensack Meridian Health
Carrier Clinic, Belle Mead, New Jersey.  The participants were diagnosed by medical professionals
at the clinic.  The 74 participants comprised the following four categories: 25 healthy participants 
(no mental health disorder), 23 participants with bipolar disorder, 10 participants with major 
depressive disorder, and 16 participants with schizoaffective disorder.  The experimental procedure 
for data collection and analysis was approved by the Institutional Review Board of Princeton 
University.  The physiological signals of the participants were captured by a commercially-available
Empatica E4 smartwatch \cite{e4-connect} and a Samsung Galaxy S4 smartphone, as shown in 
Fig.~\ref{fig:devices}.  We summarize all the data types collected in this study in 
Table~\ref{tab:data-types}.  The physiological signals are derived from WMSs embedded in the 
smartwatch. They include GSR that measures sympathetic nervous system arousal, IBI that indicates
the heart rate, ST that provides skin temperature readings, and $3$-axis accelerometer (Acc-W) that 
measures acceleration in the $X$, $Y$, and $Z$ directions.  Ambient and motion information is 
captured using sensors in the smartphone that include ambient temperature (Temp), gravity (Grav), 
acceleration (Acc-P), and angular velocity (Vel).  
It is worth mentioning that the acceleration sensors in the smartphone and smartwatch have different sampling rates, and capture different motion information. 

Before data collection, all participants are informed about the experiment and are asked
to sign a consent form. The data collection setup consists of placing the Empatica E4
smartwatch on the wrist of the participant's non-dominant hand and placing the Samsung
Galaxy S4 smartphone in the opposite front pocket. Data collection lasts around 1.5 hours, during
which time the participant is allowed to freely move around in the room with their on-body devices. 
During this time, the smartwatch and smartphone continuously record and store
the physiological signals and ambient/motion information.  At the end of
the data collection period, we remove the smartwatch from the patient's wrist and the smartphone from
their pocket. We use the Empatica E4 Connect portal for smartwatch data retrieval. 
We use a private Android application to download the smartphone data streams. All of the
recorded data are timestamped at the time of sampling.

Next, we preprocess the dataset for use in DNN training.  We first synchronize the smartwatch and
smartphone data streams for each participant.  This is necessary since the WMS data streams may vary 
in their start times and frequencies.  Then, we divide the data for each participant into 15-second 
windows.  Each 15-second window of the combined smartwatch/smartphone data constitutes one data 
instance.  There is no time overlap between data instances.  To obtain each data instance, we flatten 
and concatenate the data within the same time window from both the smartwatch and smartphone. This 
results in a feature space of dimension 2325. The smartwatch (smartphone) contributes 1575 
(750) features. All the smartphone sensors have a sampling rate of 5Hz. In addition, the smartwatch 
sensors include one data stream at 32Hz, two data streams at 4Hz, and one data stream at 1Hz. 

\begin{table}[]
\caption{Data types collected in the MHDeep framework}
\label{tab:data-types}
% \resizebox{\textwidth}{!}{
\centering
\begin{tabular}{lcl}
\toprule
Data type       & Sampling rate (Hz) & Data source   \\ 
\toprule
Galvanic skin response ($\mu$S) & 4 & Smartwatch   \\
Skin temperature ($^\circ C$) & 4 & Smartwatch   \\ 
Inter-beat interval ($ms$) & 1 & Smartwatch   \\ 
Acceleration ($x, y, z$) & 32 & Smartwatch \\
\midrule
Ambient temperature (\%)& 5 & Smartphone \\
Gravity ($x, y, z$) & 5 & Smartphone\\
Acceleration ($x, y, z$) & 5 & Smartphone\\
Angular velocity ($x, y, z$) & 5 & Smartphone\\
\bottomrule
\end{tabular}
\end{table}

For each classification task, since the number of individuals in each of the four categories is 
small, we created three different data partitions for evaluation.  The data instances extracted from 
the individuals in each of the four groups (healthy, schizoaffective, depressive, and bipolar) were 
divided into three sets: training, validation, and test.  To evaluate the models on different unseen 
patients, data instances included in the training, validation, and test sets came from
different individuals, i.e., no individual contributed data to more than one of these sets.
Among the healthy participants, for each of the three data partitions, data instances from 15 
individuals (60\% of the healthy participants) are selected for the training set, from 5 individuals 
(20\% of the healthy participants) for the validation set, and from the remaining 5 individuals 
(20\% of the healthy participants) for the test set.  For individuals with bipolar disorder, the 
training, validation, and test sets contain data instances from 13, 5, and 5 participants, 
respectively. Among the participants who had major depressive disorder, data instances from 6 
participants are selected for the training set and from 2 participants each for the validation and 
test sets.  For individuals with schizoaffective disorder, the training, validation, and test sets 
include data instances from 10, 3, and 3 participants, respectively. 

We create the final dataset for each binary classification task (healthy vs. the mental health
disroder) by combining the training, validation, and test sets of the two classes involved in that 
task. We use SMOTE \cite{chawla2002smote} to up-sample data instances from the minority class.  
Table~\ref{tab:data-instances} shows the number of instances for each of the classification tasks for 
all three data partitions. 

\begin{table*}[]
\caption{Details of various datasets (MDD: major depressive disorder).}
\label{tab:data-instances}
% \resizebox{\textwidth}{!}{
\centering
\begin{tabular}{l|c|ccc}
\toprule
Classification task & Data partition  &      Training instances (\#individuals) & Validation instances (\#individuals)     & Test set  (\#individuals)  \\ 
\toprule
Healthy vs. Bipolar & 1 & 14828 (28) & 3582 (10) & 3754 (10)  \\
Healthy vs. MDD & 1 &  14828 (21) & 2414 (7) & 2515 (7) \\ 
Healthy vs. Schizo. & 1 & 14828 (25) & 2789 (8) & 3047 (8) \\ 
\midrule
Healthy vs. Bipolar & 2 & 13330 (28) &  3922 (10) & 3582 (10)  \\
Healthy vs. MDD & 2 & 13330 (21) & 3266 (7) & 2414 (7)   \\ 
Healthy vs. Schizo. & 2 &  13330 (25) & 3773 (8) & 2789 (8)   \\ 
\midrule
Healthy vs. Bipolar & 3  & 12054 (28) & 4102 (10) & 3922 (10) \\
Healthy vs. MDD & 3  & 12054 (21) & 3088 (7) & 3266 (7) \\ 
Healthy vs. Schizo. & 3 & 12054 (25) & 3522 (8) & 3773 (8)  \\ 
\bottomrule
\end{tabular}
\end{table*}

\subsection{MHDeep DNN synthesis}
\label{sect:dnn-synthesis}
\begin{figure}[!ht]
    \centering
    \includegraphics[scale= 0.45]{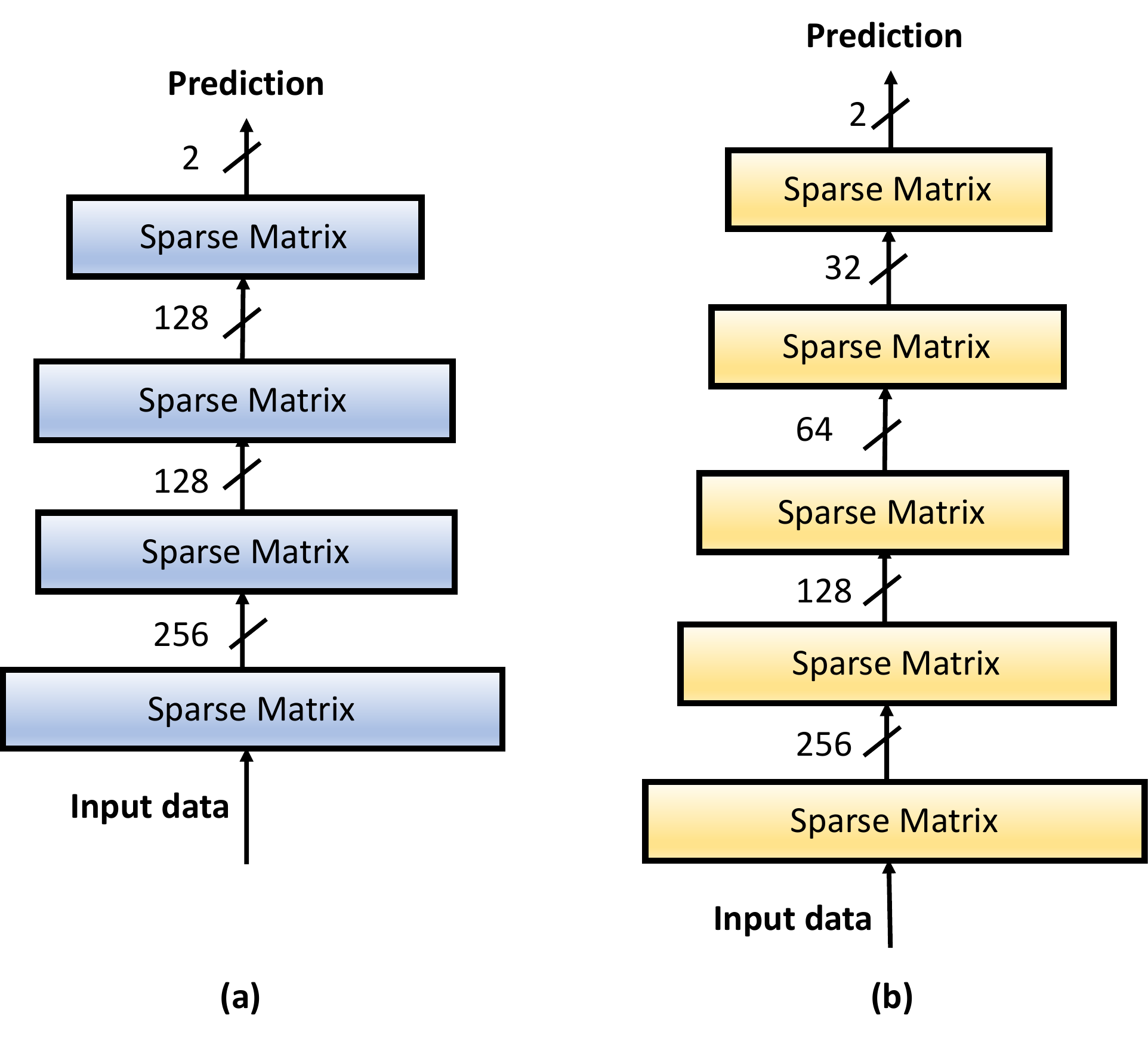}
    \caption{Architecture of MHDeep DNNs for: (a) healthy vs. major depressive disorder and 
healthy vs. schizoaffective disorder, and (b) healthy vs. bipolar disorder.}
\label{fig:architectures}
\end{figure}

Fig.~\ref{fig:architectures} shows the DNN architectures used in the MHDeep framework. 
The architectures receive the input data at the bottom and make their diagnostic decisions at the 
top. For the healthy vs. major depressive disorder and healthy vs. schizoaffective disorder
binary classification tasks, the DNN architecture has four layers with a width of 256, 128,
128, and 2, respectively. For the healthy vs. bipolar disorder binary classification task, we
use a DNN architecture with five layers with a width of 256, 128, 64, 32, and 2, respectively. 
We selected these DNN architectures by verifying the performance of various DNNs (with different 
numbers of layers and number of neurons per layer) on the validation set and picking the best
performing one.  These architectures are initially fully-connected.

We then subject the fully-connected DNNs to three sequential steps: (i) synthetic data generation 
to mimic the distribution of the real training data, (ii) pre-training of the DNN architectures with 
the synthetic data, and (iii) grow-and-prune DNN synthesis to reduce the redundancy of the DNN model 
while improving its performance. Next, we discuss each step in more detail. 

\begin{figure*}[!ht]
    \centering
    \includegraphics[scale=0.5]{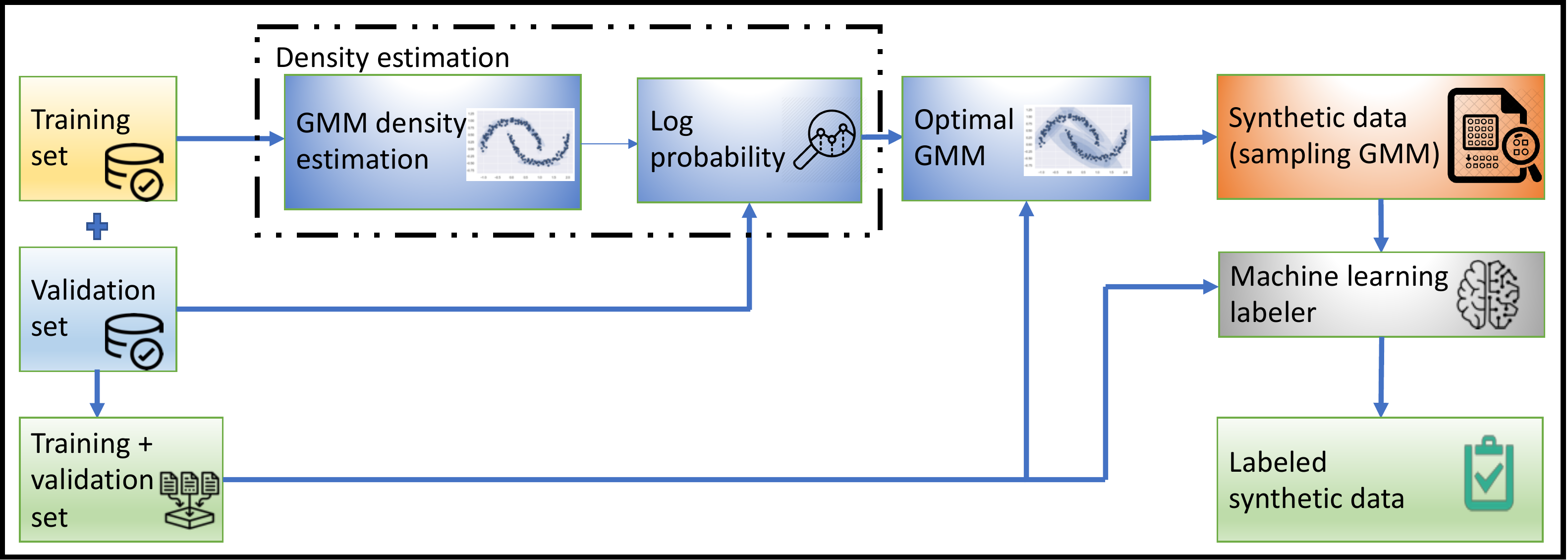}
    \caption{Schematic diagram of the MHDeep synthetic data generation process.}
\label{fig:syn-data}
\end{figure*}

\begin{enumerate}
\item \textbf{Synthetic data generation}: In this step, we generate a synthetic dataset
that mimics the probability distribution of the real training dataset. Fig.~\ref{fig:syn-data} 
illustrates the synthetic data generation process. This approach was first presented in 
\cite{hassantabar2020tutor} to alleviate the need for large datasets to train DNN architectures. 
We first use GMM to estimate the density of the training dataset.  We optimize the number of 
mixtures in the GMM by monitoring the likelihood of validation data. We choose the number of 
components that maximizes the following criterion:
    
$$
N^{*} = \underset{N}{\arg\max} \left( \texttt{GMM}_{N}(x).\texttt{score}(X_{validation}) \right)
$$
Finally, we train the optimal GMM model with $N^{*}$ mixtures on the combination of the
training and validation datasets. By sampling this model, we are able to generate the synthetic data:
$$
X^{*} = GMM_{N^{*}}(X_{total}).sample()
$$
For our experiments, we generate 100,000 samples as synthetic data. 
The final step is labeling of the synthetic dataset. We use a traditional machine learning model 
for this purpose.  We evaluate various models, e.g., the support vector machine and random forest 
models based on different splitting criteria (such as Gini index and entropy), and different depth 
limits on the decision trees, on the validation set.  The model with the highest accuracy is 
used to label the synthetic data. 
    
\item \textbf{DNN pre-training}: 
In this step, we use the labeled synthetic data to obtain a prior on the weights of the DNN 
architecture by pre-training them.  The intuition behind this step is that pre-training the DNN 
provides a suitable inductive bias to the DNN. As a result, we can commence on the final training 
stage with a better weight initialization.  Therefore, it alleviates the need for large training 
datasets. 
    
\item \textbf{Grow-and-prune DNN synthesis}: 
MHDeep uses a grow-and-prune DNN synthesis paradigm to train the models. 
Algorithm~\ref{alg:grow-and-prune} summarizes this process.  It uses a mask-based approach. 
For each weight matrix, there is an associated binary mask of the same size that is used to disregard 
dormant connections in the architecture.  It applies magnitude-based pruning and full growth to
fully-connected DNNs iteratively.  For magnitude-based pruning, a hyperparameter $\alpha$ is used to
depict the pruning ratio. We prune a connection if and only if its weight is in the lowest 
$\alpha*100$ percent of the weights in its associated layer.  Finally, for the pruned connections, we 
set the weight and its binary mask both to $0$.  Since connection pruning is an iterative process, we 
retrain the network to recover its performance after each pruning iteration.  In our
experiments, after each architecture changing operation, we train the DNN for 20 epochs. In addition, 
we set the number of iterations to $5$. 
    
\begin{algorithm}[h]
    \caption{Grow-and-prune synthesis}
    \label{alg:grow-and-prune}
    \begin{algorithmic}[l]
        \REQUIRE Pre-trained DNN architecture; iteration number $numIterations$;
        weight matrix $W \in R^{M \times N}$; mask matrix $Mask$ of the same dimension as 
        the weight matrix; $\alpha$: pruning ratio
        \STATE best-validation-acc = 0
        \FORALL {layers in the DNN}
        \STATE $t = (\alpha \times MN) ^{th}$ largest element in $\left|W\right|$
        \FORALL {$w_{ij}$}
        \IF{$\left| w_{ij} \right| < t$}
        \STATE {$Mask_{ij} = 0$}
        \ENDIF
        \ENDFOR
        \STATE $W$ = $W \otimes Mask$ 
        \ENDFOR
        \STATE Train DNN for given \#epochs
        \STATE validation-acc = evaluate DNN on validation set
        \IF{validation-acc \textgreater best-validation-acc}
        \STATE best-validation-acc = validation-acc
        \STATE Save the DNN
        \ENDIF
        
        \FOR{$numIterations$}
        
        \FORALL {layers in the DNN}
        \STATE $Mask_{[1:M, 1:N]} = 1 $
        \ENDFOR
        
        \FORALL {layers in the DNN}
        \STATE $t = (\alpha \times MN) ^{th}$ largest element in $\left|W\right|$
        \FORALL {$w_{ij}$}
        \IF{$\left| w_{ij} \right| < t$}
        \STATE {$Mask_{ij} = 0$}
        \ENDIF
        \ENDFOR
        \STATE $W$ = $W \otimes Mask$ 
        \ENDFOR
        \STATE Train DNN for given \#epochs
        \STATE validation-acc = evaluate DNN on validation set
        \IF{validation-acc \textgreater  best-validation-acc}
        \STATE best-validation-acc = validation-acc
        \STATE Save the DNN
        \ENDIF
        
        \ENDFOR
        \ENSURE Best architecture with the weight and mask matrices
    \end{algorithmic}
\end{algorithm}
    
\end{enumerate}

\subsection{MHDeep inference process}
\label{sect:inference}
The trained DNN models can be used for diagnosis or daily monitoring of the mental state of the
user based on collection of physiological signals and ambient information during the day. 
The collected data streams are processed based on the step explained in 
Section~\ref{sect:data-collection}.  We feed the processed data to the MHDeep DNN that predicts
the mental health condition of the user.  When the model predicts that the user is experiencing an
episode of mental health disorder, this information can be sent to a physician for early treatment.

\section{Implementation Details}
\label{sect:implementation}
We have implemented the data processing and preparation parts of the MHDeep framework in Python and
the MHDeep DNN synthesis framework in PyTorch. We use the Nvidia Tesla P$100$ data center accelerator
for DNN training. We use the cuDNN library to accelerate GPU processing.  For training, we use 
a stochastic gradient descent (SGD) optimizer, with a learning rate of $5$e-$4$ and a batch size of 
$256$.  We use 100,000 synthetic data instances to pre-train the network architecture.  
In the grow-and-prune synthesis phase, we train the network for $20$ epochs each time the 
architecture changes. We use an SGD optimizer, with an initialized learning rate of $1$e-$4$
that we halve in each succeeding iteration.  We apply network-changing operations over five 
iterations.  

\section{Evaluation}
\label{sect:evaluation}
In this section, we analyze the performance of MHDeep DNN models for diagnosing three 
mental health disorders. This entails three binary classifications: (i) schizoaffective disorder
vs. healthy individuals, (ii) major depressive disorder vs. healthy individuals, and (iii) 
bipolar disorder vs. healthy individuals. For each classification task, we use three different data 
partitions, each partition with data instances obtained from different individuals in the training, 
validation, and test sets.

The MHDeep DNN models are evaluated with four different metrics: test accuracy, false positive rate 
(FPR), false negative rate (FNR), and F1 score.  Accuracy measures overall classification 
performance. It is simply the ratio of all the correct predictions on the test data instances and 
the total number of such instances. FPR and FNR measure how often healthy individuals are
declared to have the corresponding mental health condition and {\em vice versa}, respectively.

First, we report the performance of the MHDeep DNN models in detecting each of the three mental 
health disorders at the data instance level.  Next, we evaluate the accuracy of the models in 
detecting mental health disorders at the patient level. 

\subsection{MHDeep performance evaluation at the data instance level}
We first analyze the performance of the three binary classifiers.  We begin by training DNN models 
on features obtained from subsets of the eight data categories presented in 
Table~\ref{tab:data-types}.  We analyze all the subsets of the eight data categories and report 
results for the top models.  Since there are eight data categories, there are 256 subsets, with
one being the null subset.  We evaluated the remaining 255 subsets.  This helps 
distinguish the impact of each data category and to find the most effective combination 
of categories for each classification task. 

Table~\ref{tab:hvs-results} shows the results for classification between healthy and schizoaffective 
data instances. The best data category subset, in this case, achieves an average test accuracy of 
90.4\%. We also report test accuracy, FPR, FNR, and F1 score for each of the three data partitions. 
The model reaches the highest test accuracy of 93.3\% on the second data partition.  Furthermore, 
for the healthy instances, the top model achieves a low average FPR of 6.5\%, demonstrating its 
effectiveness in avoiding false alarms.  For the schizoaffective instances, the model achieves
an average FNR of 16.9\%, indicating reasonable effectiveness in raising alarms when schizoaffective 
disorder does occur. We also report the number of parameters (\#params) and floating-point 
operations (FLOPs) required for each model.  We also compare \#params and FLOPs of the models with 
those of the fully-connected baselines. As can be seen, using the grow-and-prune DNN synthesis 
approach enables us to reduce both \#params and FLOPs, leading to a reduction in memory and 
computational requirements. 

We present the results for classification between healthy and major depressive disorder instances in 
Table~\ref{tab:hvd-results}. The data category subset with the best performance achieves an average 
test accuracy of 87.3\%. This model achieves the highest accuracy of 91.2\% on the second data 
partition. It achieves an average FPR (FNR) of 6.8\% (29.3\%).  

Table~\ref{tab:hvb-results} presents the results for classification between healthy and bipolar 
disorder instances. In this case, the model trained on the best data category subset achieves an 
average test accuracy of 82.4\%, with an FPR (FNR) of 16.7\% (20.7\%). 

\begin{table*}[]
\caption{Test accuracy, FPR, FNRs, and F1 score (all in \%) for top data categories for
classification between healthy and schizoaffective disorder data instances}
\label{tab:hvs-results}
\centering
\begin{tabular}{lccccccc}
\toprule
Data category & Data partition & \#Params (compression) & FLOPs (compression) & Acc. & FPR & FNR & F1 Score\\
\midrule
(Acc-P, Temp, Vel, Acc-W, GSR, IBI) & 1 & 275.0k (2.1$\times$) & 549.5k (2.0$\times$) & 91.2 & 11.4 & 5.8 & 89.5\\
 & 2 & 300.0k (1.9$\times$) & 599.5k (1.9$\times$) & 81.4 & 6.6 & 37.8 & 72.0\\
 & 3 & 275.0k (2.1$\times$) & 549.5k (2.0$\times$) & 87.3 & 2.4 & 34.1 &  84.2\\
\midrule
Average &  & & &  86.6 & 6.8 & 25.9 & 81.9\\
\midrule
\bottomrule
(Acc-P, Temp, Vel, Acc-W, GSR) & 1 & 200.0k (2.8$\times$) & 399.5k (2.8$\times$) & 90.5 & 13.0 & 4.4 & 89.3\\
 & 2 & 300.0k (1.9$\times$) & 599.5k (1.8$\times$) & 93.3 & 4.0 & 13.3 &  89.8\\
 & 3 & 250.0k (2.3$\times$) & 499.5k (2.2$\times$) & 87.5 & 2.5 & 32.9 &  77.9\\
\midrule
Average &  & & & 90.4 & 6.5 & 16.9  & 85.7\\
\midrule
\bottomrule
(Acc-P, Temp, Grav, Vel, Acc-W, GSR, IBI)  & 1 & 300.0k (2.1$\times$) & 599.5k (2.0$\times$) & 88.3 & 14.4 & 7.7 & 86.8\\
 & 2 & 350.0k (1.8$\times$) & 699.5k (1.8$\times$) & 82.6 & 5.9 & 45.7 &  66.3\\
 & 3 & 300.0k (2.1$\times$) & 599.5k (2.0$\times$) & 88.7 & 3.9 & 26.4 &  81.1\\
\midrule
Average &  & & &  86.5 & 8.1 & 26.6 & 78.1\\
\midrule
\bottomrule
\end{tabular}
\end{table*}

\begin{table*}[]
\caption{Test accuracy, FPR, FNRs, and F1 score (all in \%) for top data categories for classification between healthy and major depressive disorder data instances}
\label{tab:hvd-results}
\centering
\begin{tabular}{lccccccc}
\toprule
Data category & Data partition & \#Params (compression) & FLOPs (compression) &  Acc. & FPR & FNR & F1 Score\\
\midrule
(Temp, Grav, Vel, GSR) & 1 & 75.0k (2.7$\times$) & 149.5k (2.4$\times$) & 89.0 & 4.6 & 26.7 & 79.4\\
 & 2 & 120.0k (1.7$\times$) & 239.5k (1.5$\times$) & 90.5 & 4.5 & 21.7 & 82.7\\
 & 3 & 150.0k (1.3$\times$) & 299.5k (1.2$\times$) & 81.7 & 12.7 & 37.9 &  60.2\\
\midrule
Average &  & & & 87.1 & 7.3 & 28.8 & 74.1\\
\midrule
\bottomrule
(Acc-P, Grav, Vel, GSR) & 1 & 120.0k (2.0$\times$) & 239.5k (1.8$\times$) &  88.2 & 6.5 & 24.8 & 78.7\\
 & 2 & 145.0k (1.6$\times$) & 289.5k (1.5$\times$) & 91.2 & 1.9 & 25.7 &  83.0\\
 & 3 & 185.0k (1.3$\times$) & 369.5k (1.2$\times$) & 82.4 & 11.9 & 37.5 &  61.3\\
\midrule
Average &  & & & 87.3 & 6.8 & 29.3  & 74.3\\
\midrule
\bottomrule
\end{tabular}
\end{table*}

\begin{table*}[]
\caption{Test accuracy, FPR, FNRs, and F1 score (all in \%) for top data categories for
classification between healthy and bipolar disorder data instances}
\label{tab:hvb-results}
\centering
\begin{tabular}{lccccccc}
\toprule
Data category & Data partition & \#Params (compression) & FLOPs (compression) & Acc. & FPR & FNR & F1 Score\\
\midrule
(Acc-P, Temp, Grav, Vel, Acc-W, IBI, ST) & 1 & 500.0k (1.2$\times$) & 999.5k (1.2$\times$) & 76.1 & 47.1 & 2.8 & 81.0\\
 & 2 & 480.0k (1.3$\times$) & 959.5k (1.3$\times$) &  81.4 & 1.0 & 34.3 & 78.9\\
 & 3 & 400.0k (1.6$\times$) & 799.5k (1.5$\times$) & 89.8 & 2.1 & 25.1 &  83.8\\
\midrule
Average &  & & &  82.4 & 16.7 & 20.7 & 81.2\\
\midrule
\bottomrule
(Temp, Grav, Vel, Acc-W, GSR, IBI) & 1 & 490.0k (1.2$\times$) & 979.5k (1.1$\times$) & 75.6 & 47.1 & 3.8 & 80.5\\
 & 2 & 450.0k (1.3$\times$) & 899.5k (1.2$\times$) & 81.6 & 0.9 & 34.2 &  79.0\\
 & 3 & 380.0k (1.5$\times$) & 759.5k (1.5$\times$) & 87.0 & 2.1 & 33.0 &  78.4\\
\midrule
Average &  & & &  $81.4$ & $16.7$ & $23.7$  & $79.3$\\
\midrule
\bottomrule
\end{tabular}
\end{table*}

\subsection{MHDeep performance evaluation at the patient level}
\begin{figure*}[!ht]
    \centering
    \includegraphics[scale=0.5]{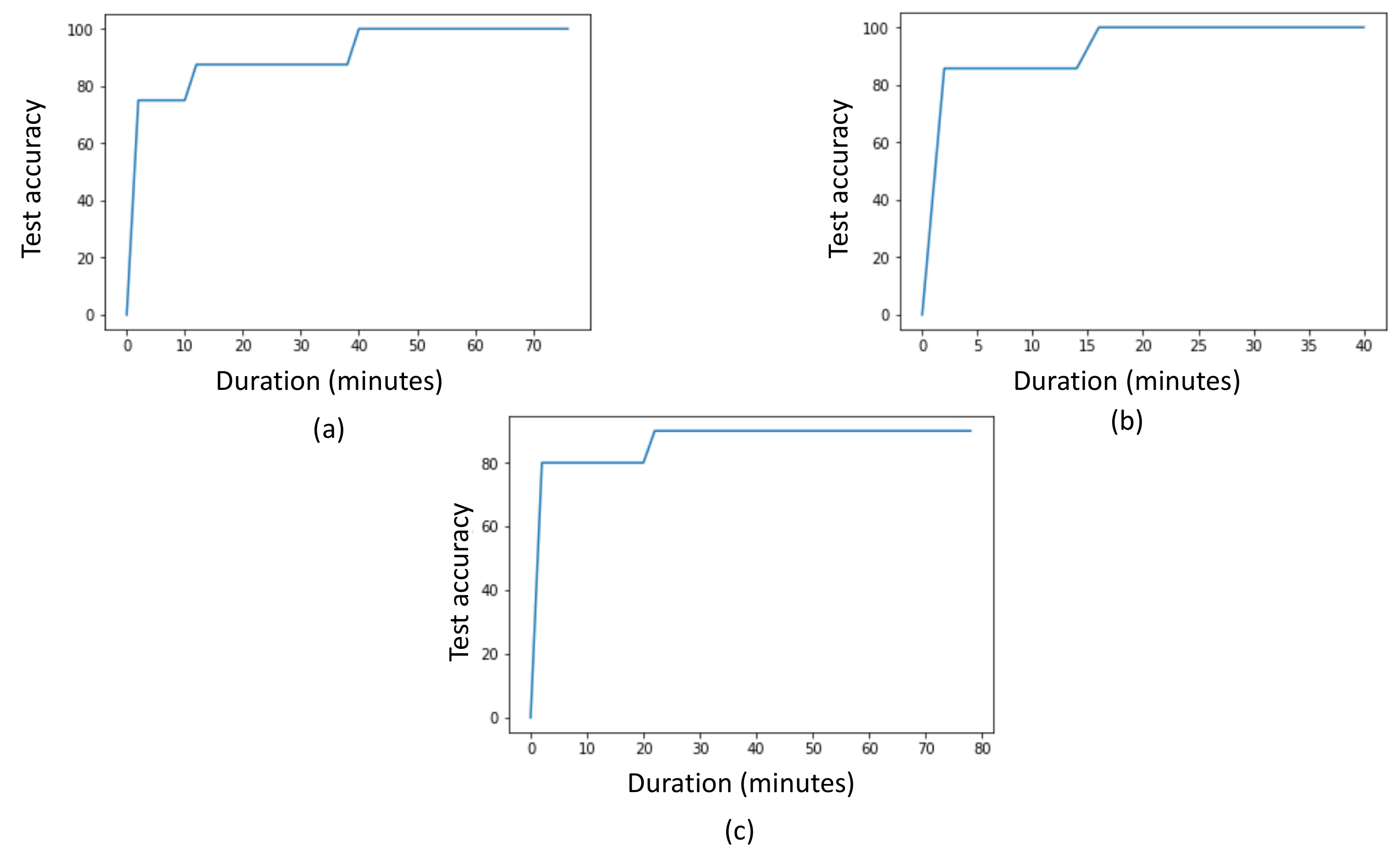}
    \caption{Patient-level test accuracy vs.~duration of data needed for classification between 
(a) healthy and schizoaffective disorder individuals, (b) healthy and major depressive disorder 
individuals, and (c) healthy and bipolar disorder individuals.}
\label{fig:patient-acc}
\end{figure*}

Next, we show patient-level diagnostic test accuracy.  We use the most accurate model from among the 
models discussed above for each classification task.  Fig.~\ref{fig:patient-acc} shows the results. 
In these graphs, we plot patient-level test accuracy vs.~the duration of data needed for inference. 
Prediction is performed for each patient by simply taking the majority of the predicted labels for
each 15s data instance in the given data duration.  We step up the data duration size by 2
minutes each time.  Thus, we add eight data instances in each 2-minute window. 
As we can see, the models reach 100\% test accuracy after a certain point for distinguishing 
healthy individuals from those with schizoaffective and major depressive disorders.
In addition, the best model for classification between healthy and bipolar disorder individuals 
reaches 90.0\% patient-level accuracy.  Table \ref{tab:duration} shows the minimum data duration needed 
to reach the saturation accuracy.  The durations are 40, 16, and 22 minutes for healthy vs.
schizoaffective disorder, healthy vs. major depressive disorder, and healthy vs. bipolar disorder 
classifications, respectively. 

\begin{table}[]
\caption{Minimum inference data duration (in minutes) needed to reach saturation patient-level 
accuracy (in \%) for each classification task}
\label{tab:duration}
\centering
\begin{tabular}{lcc}
\toprule
Classification & Time & Saturation accuracy \\
\midrule
Healthy vs. Schizoaffective disorder& 40 &  100 \\
Healthy vs. Major depressive disorder & 16 & 100\\
Healthy vs. Bipolar disorder & 22 & 90.0 \\
\bottomrule
\end{tabular}
\end{table}

\section{Discussion}
\label{sect:discussion}
MHDeep combines efficient neural networks with commercially available WMSs to diagnose various mental 
health disorders.  Although several works address mental health problem detection using machine 
learning, to our knowledge, MHDeep is the only solution that focuses on an easy-to-use system that 
can monitor the daily mental health state of the user.  The diagnostic decisions can be sent to a 
health server from where medical professionals can access the information.  This can enable
them to quickly intervene during severe episodes of the disorder. 

We demonstrated the diagnostic effectiveness of MHDeep for three different mental health disorders. 
However, this work can be generalized to other types of mental health disorders as well.  We hope it 
will encourage researchers to start collecting WMS data from individuals across a diverse set of 
challenging diagnostic tasks.  Bypassing the manual feature engineering step through the use of 
efficient DNNs enables easy scalability of this approach to many other disease domains.  It can also 
be used to predict the progress of a disease based on longitudinal WMS data collected in the
training stage. 

Diagnosis of mental health disorders is often based on patient's self-report and answers to a 
questionnaire designed to detect each disorder. In the future, we can improve the performance of 
MHDeep by adding a specifically designed questionnaire to the data categories. In addition, adding 
features based on other WMSs such as blood pressure may also help enhance model performance. 

\section{Conclusion}
\label{sect:conclusion}
In this article, we proposed a framework called MHDeep that combines data obtained from commercially 
available WMSs with the knowledge distillation power of DNNs for continuous and pervasive diagnosis 
of three main mental health disorders: schizoaffective, major depressive, and bipolar. MHDeep uses a 
synthetic data generation module to address the lack of large datasets.  We trained the DNN models by 
using iterative network growth and pruning to learn both the weights and architecture during the 
training process.  We evaluated MHDeep based on data collected from 74 individuals. It achieves 
patient-level accuracy of 100\%, 100\%, and 90.0\%, using 40, 16, and 22 minutes of data
collected in the inference stage, for classification between healthy and schizoaffective disorder 
individuals, healthy and major depressive disorder individuals, and healthy and bipolar disorder 
individuals, respectively.  The MHDeep models were also shown to be computationally efficient.
Thus, MHDeep can be employed for pervasive diagnosis and daily monitoring while offering high 
computational efficiency and accuracy. 

\noindent
{\bf Acknowledgments:} We would like to thanks the nurses and physicians in the Acute Care Unit East and West at the Hackensack Meridian Health Carrier Clinic for smartwatch/smartphone based data collection from various patient cohorts and
providing patient labels. Special thanks to Donald J. Parker, CEO of Carrier Clinic, for recognizing 
the promise of such a study and facilitating data collection.
\bibliographystyle{IEEEtran}
\bibliography{MHDeep}
% comments on bibliography
% - 19, 32, should have in PRoc...

\vspace{-1.0cm}
\begin{IEEEbiography}[
{\includegraphics[width=1.0in,height=1.1in,clip,keepaspectratio]{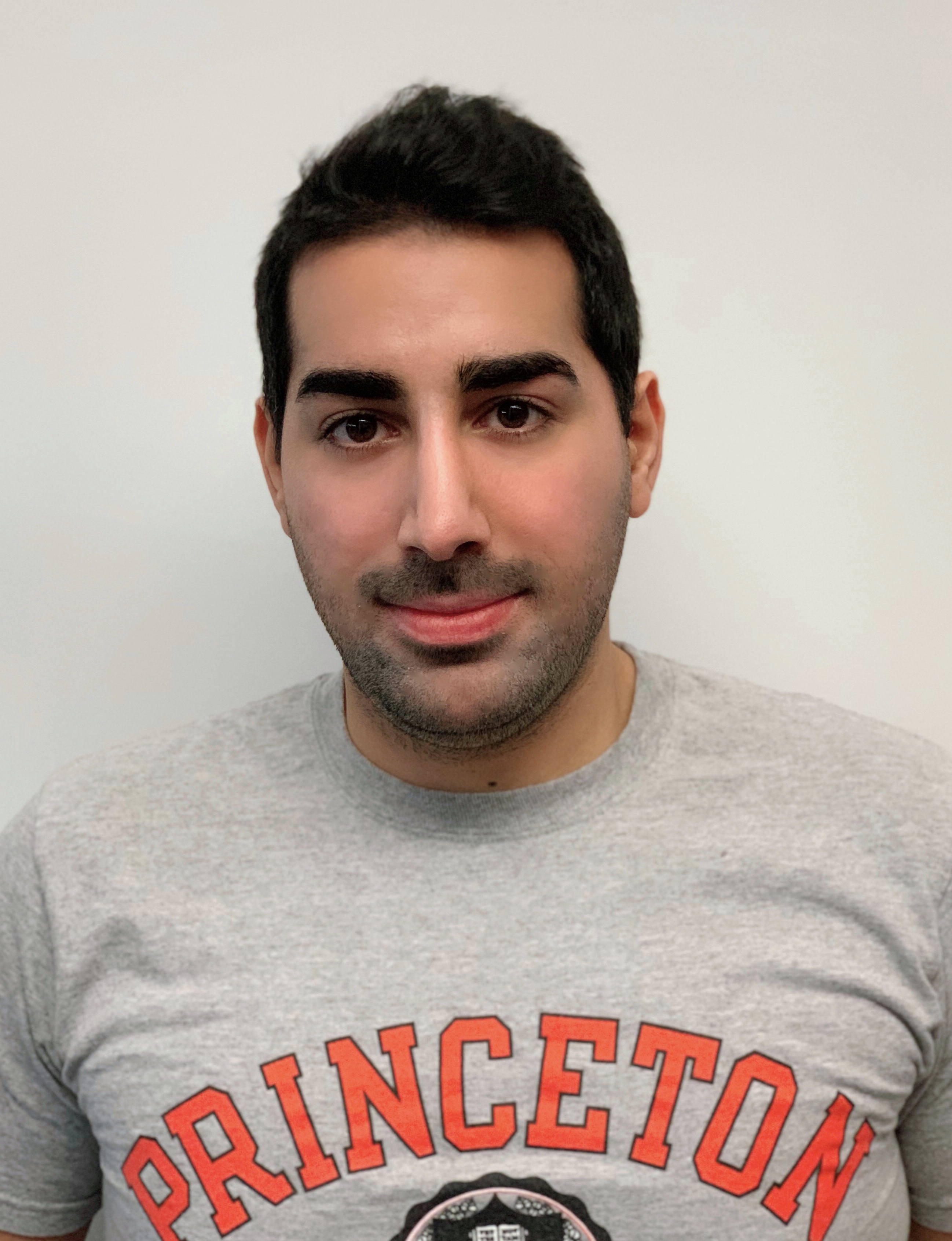}}]
{Shayan Hassantabar} received his B.S. degree in Electrical Engineering, Digital Systems focus, from Sharif University of Technology, Iran. He also received his M.Math. degree in Computer Science from University of Waterloo, Canada, and his M.A. degree in Electrical and Computer Engineering from Princeton University. He is pursuing the Ph.D. degree in Electrical and Computer Engineering at Princeton University. His research interests include automated neural network architecture synthesis, neural network compression, and smart healthcare.
\end{IEEEbiography}

\vspace{-1.0cm}
\begin{IEEEbiography}[
{\includegraphics[width=1.0in,height=1.1in,clip,keepaspectratio]{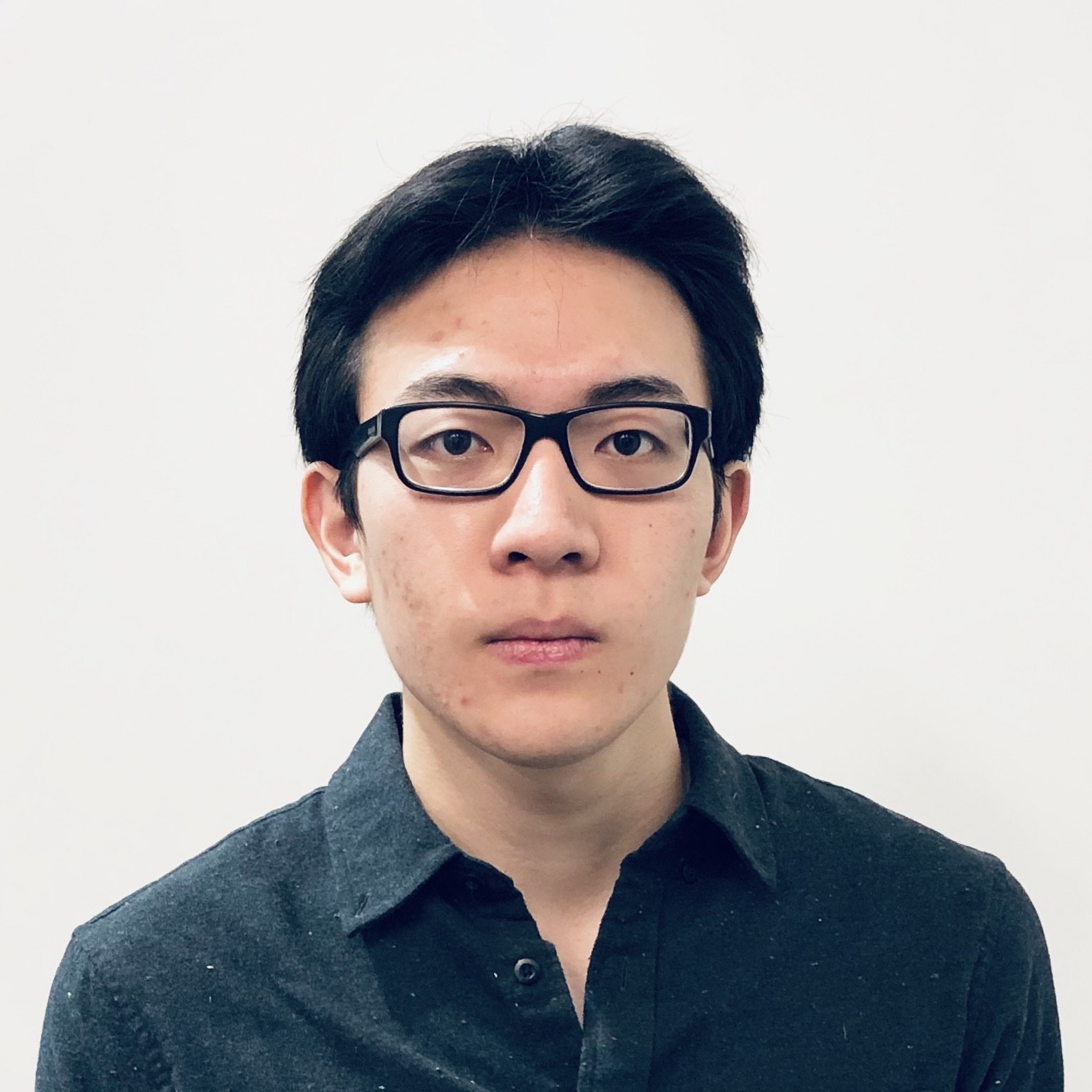}}]
{Joe Zhang} received his B.S.E. from Princeton University in 2020. He is now pursuing a Ph.D. in Electrical Engineering at Stanford University, supported by a Stanford Graduate Fellowship. At Princeton, he was the recipient of the Shapiro Prize for Academic Excellence and the Hisashi Kobayashi Prize. He has worked on deep learning for healthcare applications and his current research at Stanford focuses on developing robust protocols and systems for blockchain technologies.
\end{IEEEbiography}

\vspace{-1.0cm}
\begin{IEEEbiography}[
{\includegraphics[width=1.0in,height=1.1in,clip,keepaspectratio]{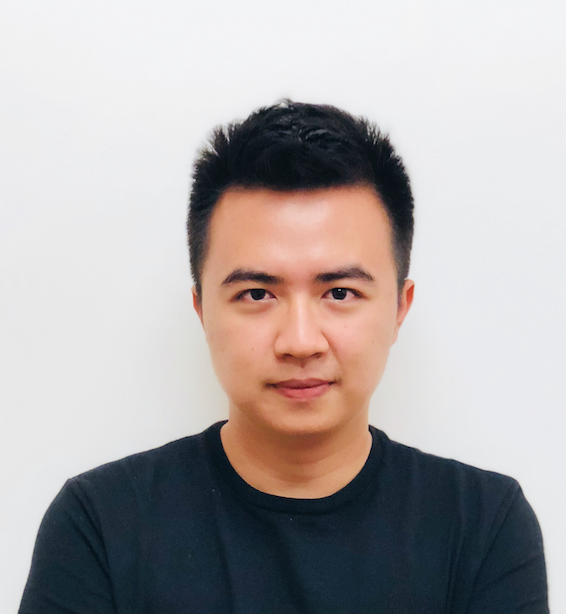}}]
{Hongxu Yin} received his Ph.D. from Princeton University in 2020. He received his B.Eng. degree from Nanyang Technological University, Singapore,
in 2015. He is now a Research Scientist with NVIDIA Research. He is a recipient of Princeton Yan Huo 94* Graduate Fellowship, Princeton Natural Sciences and Engineering Fellowship, Defense Science \& Technology Agency gold medal, and Thomson Asia Pacific Holdings gold medal. His research focuses on efficient deep neural networks, data-free and hardware-guided model compression, and efficient inference for healthcare applications.
\end{IEEEbiography}

\vspace{-1.0cm}
\begin{IEEEbiography}[
{\includegraphics[width=1.0in,height=1.1in,clip,keepaspectratio]{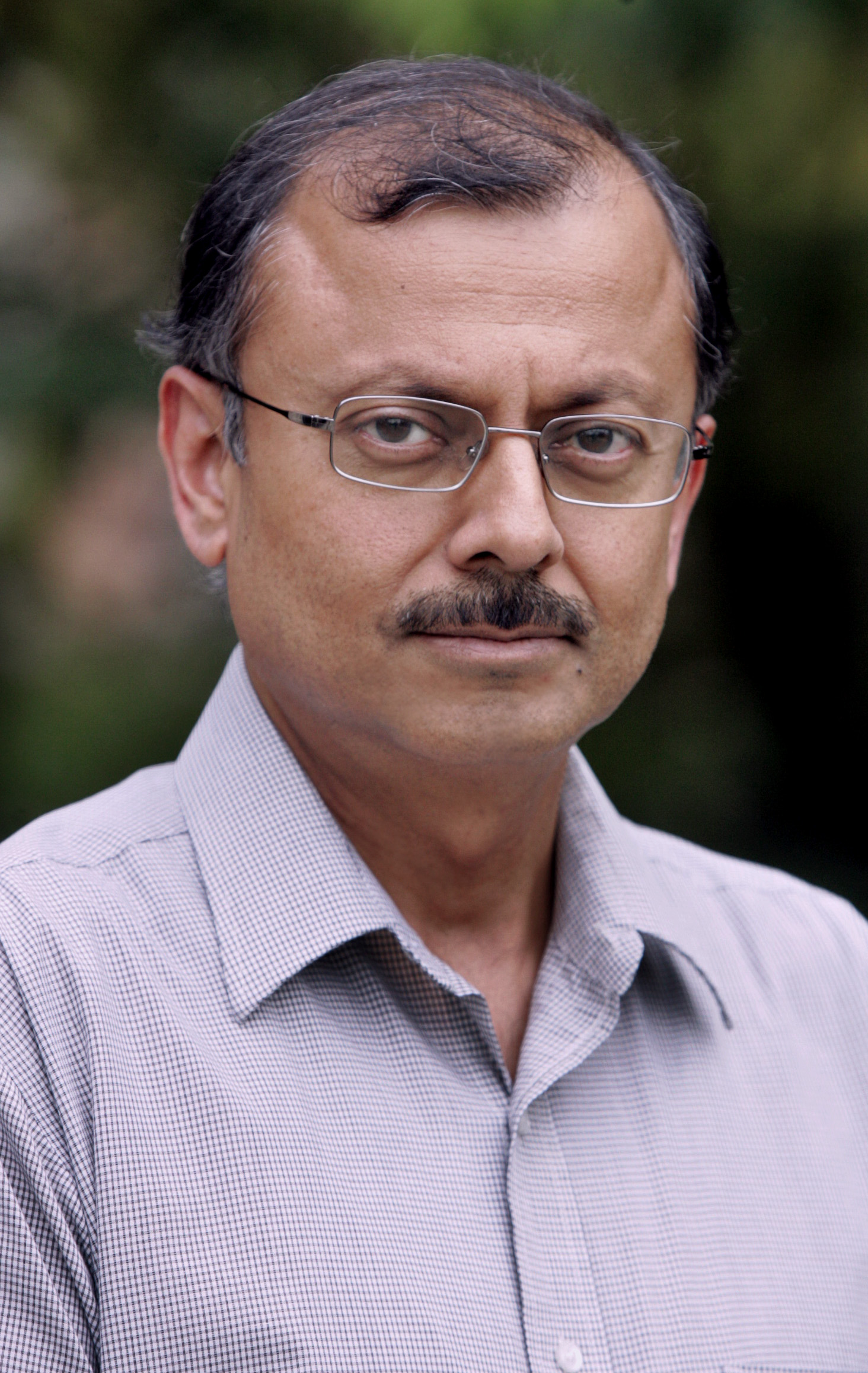}}]
{Niraj K. Jha} received his B.Tech. degree in Electronics and Electrical Communication Engineering from Indian Institute of Technology, Kharagpur, India in 1981 and Ph.D. degree in Electrical Engineering from University of Illinois at Urbana-Champaign, IL in 1985. He has been a faculty member of the Department of Electrical Engineering, Princeton University, since 1987. He is a Fellow of IEEE and ACM, and was given the Distinguished Alumnus Award by I.I.T., Kharagpur. He has received the Princeton Graduate Mentoring Award.

He has served as the Editor-in-Chief of IEEE Transactions on VLSI Systems and an Associate Editor of several other journals. He has co-authored five widely used books. His research has won 20 best paper awards or nominations and 21 patents.  His research interests include smart healthcare, cybersecurity, machine learning, and monolithic 3D IC design. He has given several keynote speeches in the areas of nanoelectronic design/test, smart healthcare, and cybersecurity.
\end{IEEEbiography}

\end{document}